\documentclass[conference]{IEEEtran}
\IEEEoverridecommandlockouts
\usepackage{mathrsfs}
\usepackage{bm}
\usepackage{cite}
\usepackage{amsmath,amssymb,amsfonts}
\usepackage{algorithmic}
\usepackage{graphicx}
\usepackage{textcomp}
\usepackage{graphicx}
\usepackage{subfig}
\usepackage{xcolor}
\usepackage{url}
\usepackage{balance}
\usepackage{enumitem}
\def\BibTeX{{\rm B\kern-.05em{\sc i\kern-.025em b}\kern-.08em
    T\kern-.1667em\lower.7ex\hbox{E}\kern-.125emX}}
\begin{document}

\title{Lie Group Control Architectures for UAVs: a Comparison of SE$_2$(3)-Based Approaches in Simulation and Hardware\\
}

\author{Dimitria Silveria$^{1}$, Kleber Cabral$^{2}$, Peter Travis Jardine$^2$, and Sidney Givigi$^{2}$
\thanks{$^{1}$D. Silveria is with the Department of Electrical and Computer Engineering, 
        Queen's University, 19 Union St, Kingston, ON K7L 3N9
        {\tt\small dimitria.s@queensu.ca}}%
\thanks{$^{2}$  K.\ Cabral, P. T. Jardine and S.\ Givigi are with the School of Computing and the Ingenuity Labs Research Institute, 
        Queen's University, Kingston, ON K7L 3N6 Canada 
        {\tt\small kleber.cabral@queensu.ca,  p.jardine@queensu.ca, sidney.givigi@queensu.ca}}%
}
\maketitle

\begin{abstract}
This paper presents the integration and experimental validation of advanced control strategies for quadcopters based on Lie groups. We build upon recent theoretical developments on SE$_2$(3)-based controllers and introduce a novel SE$_2$(3) model predictive controller (MPC) that combines the predictive capabilities and constraint-handling of optimal control with the geometric properties of Lie group formulations. We evaluated this MPC against a state-of-the-art SE$_2$(3)-based LQR approach and obtained comparable performance in simulation. Both controllers where also deployed on the Quanser QDrone platform and compared to each other and an industry standard control architecture. Results show that the SE$_2$(3) MPC achieves superior trajectory tracking performance and robustness across a range of scenarios. This work demonstrates the practical effectiveness of Lie group-based controllers and offers comparative insights into their impact on system behaviour and real-time performance.
\end{abstract}


\section{Introduction}
Unmanned aerial vehicles (UAVs) have broad applications in areas such as defence, search and rescue, and mapping. Among them, quadcopters (also known as quadrotors) are popular due to their versatility, including easy maneuverability, vertical takeoff/landing, and o hovering, which make them appropriate to fly even in small indoor environments~\cite{al-husnawy_review-uav_2024}. 

However, their non-linear and under-actuated dynamics pose challenges for control design~\cite{Abougarair_lqr-quad_2024}. Many existing approaches focus only on stabilizing a portion of their states, usually the attitude, or rely on multiple controllers, increasing their computational cost. Furthermore, few are experimentally validated on real platforms.

A common approach for quadcopters control is the Proportional Integral Derivative (PID) controller~\cite{Lopez_pid-survey_2023}, used in~\cite{Yoon_pid-lstm_2022} with online long short-term memory (LSTM) tuning of gains for attitude stabilization during hover, and in~\cite{Liu_cascade-quad_2025} with a cascaded structure for full-state control. 
Optimal control is also commonly used.
A linear state representation of the quadrotor can be used to design a linear quadratic regulator (LQR) to stabilize the roll, pitch, yaw, and altitude~\cite{Abougarair_lqr-quad_2024}, while model predictive control (MPC) has been applied to position control of nano-quadcopters~\cite{Huu_mpc-position-crazyflie_2025}.

The approach in \cite{Cohen_2020_lqr} uses the Lie group SE$_2$(3)~\cite{barrau_2025_se23} to represent attitude, velocity, and position jointly, introducing an LQR low-level controller (based on a linearization method to obtain the state-space matrices) that stabilizes all the states simultaneously. It showed better performance than traditional LQR in simulations with large initial errors. However, it requires the entire trajectory upfront to calculate the LQR gains and practical experiments were not performed to validate the approach on real quadcopters. 

We build upon \cite{Cohen_2020_lqr} by validating its SE$_2$(3) LQR on the Quanser QDrone\footnote{\url{https://www.quanser.com/wp-content/uploads/2018/02/Qdrone-Product-Data-Sheet-v1.2.pdf}}, demonstrating improved performance over Quanser's default controllers. 
Validating controllers only in simulation does not guarantee that they work for real systems and experimental validation on real platforms of control techniques (e.g., MPC, PID, and LQR)~\cite{Okasha_experiments_2022,Sahoo_hybrid-lqr-smc_2025}. For example, \cite{khan_lqr-qdrone_2024} validates an LQR on the QDrone, but only for hovering and without comparing to Quanser's baseline.

Unlike LQR, which minimizes cost over the entire time horizon, MPC optimizes the same quadratic cost function over a moving horizon~\cite{jowski2002predictive}. Therefore, MPC does not require the whole trajectory to be known a priori and can adapt more easily to reference or environmental changes~\cite{cabral2023hierarchical}. MPC can also take into account the system's constraints, making it a suitable controller for UAVs with strict bounds on attitude, thrust, and payload.

This work makes three key contributions toward the integration and validation of advanced control strategies for UAVs. First, we present a system description of the control architecture that enables the use of SE$_2$(3)-based controllers in both simulated environments and real-world UAV platforms. Second, we propose a novel SE$_2$(3) MPC controller for 3D trajectory regulation. Third, we validate our approach through both simulation and experiments on the QDrone, showing that SE$_2$(3) MPC outperforms both the SE$_2$(3) LQR and the default QDrone PID-based controller in terms of tracking accuracy and robustness. Finally, we also provide a comparative analysis of control architectures -- cascaded PIDs (default), LQR, and MPC -- highlighting their impact on UAV perform. 

The remainder of this work is organized as follows: Section~\ref{sec:Preliminaries} introduces theoretical foundations needed for understanding the proposed methods. Section~\ref{sec:controller-model} outlines the control architecture. Section~\ref{sec:experiments} details the simulation and hardware setup., Section~\ref{sec:results} presents results and system-level impacts of each control strategy. Section~\ref{sec:conclusion} concludes the work and outlines directions for future research.

\section{Preliminaries}
\label{sec:Preliminaries}
\subsection{Lie Groups}

A general matrix Lie group, denoted by $\mathcal{G}$, is composed of $n\times n$ invertible matrices with $k$ degrees of freedom and is closed under multiplication (the interested reader is referred to \cite{Hall_quantum_2013} and \cite{martin2021lie} for a more detailed discussion). Every Lie group is associated with a Lie algebra, $\mathfrak{g}$, that represents the tangent space of the group, $T_1\mathcal{G}$, at the identity element $\bm{I}_n$. An element of the algebra can be mapped into an element of the group through the operator $\exp(\cdot$): $\mathfrak{g} \rightarrow \mathcal{G}$, and the inverse operation is performed by the operator $\log(\cdot$): $\mathcal{G} \rightarrow \mathfrak{g}$.

The hat operator $(\cdot)^{\wedge}$: $\mathbb{R}^k \rightarrow \mathfrak{g}$ maps a vector representation of an algebra element into its matrix representation, and the vee operator $(\cdot)^{\vee}$: $\mathfrak{g} \rightarrow \mathbb{R}^k$ represents the inverse operation.

SE$_2$(3)~\cite{barrau_2025_se23} (and its algebra, $\mathfrak{se}_2$(3)) is also known as the group of double direct isometries, being an extension of SE(3), including a representation of velocity, $\bm{v}$, such as

\begin{equation}\label{eq:SE_2(3)}
    \text{SE}_2\text{(3)} =
    \left\{\begin{pmatrix}
        \bm{R} & \bm{v} & \bm{p}\\
        \bm{0}_{1,3} & 1 & 0 \\
        \bm{0}_{1,3} & 0 & 1        
    \end{pmatrix}, \bm{R} \in \text{SO(3)}, \bm{v}, \bm{p} \in \mathbb{R}^3 \right\}
\end{equation}
\begin{equation}\label{eq:se_2(3)}
        \mathfrak{se}_2\text{(3)} =
        \left\{\begin{pmatrix}
        \bm{\xi_\theta}^{\wedge} & \bm{\xi}_{p} & \bm{\xi}_v\\
        \bm{0}_{1,3} & 0 & 0 \\
        \bm{0}_{1,3} & 0 & 0        
    \end{pmatrix}, \bm{\xi_\theta}, \bm{\xi}_{v}, \bm{\xi}_{p} \in \mathbb{R}^3\right\}
\end{equation}
\begin{equation} \bm{\xi} = 
        \begin{pmatrix}
        \bm{\xi}_\theta\\
        \bm{\xi}_{v}\\
        \bm{\xi}_p
    \end{pmatrix}^{\wedge} = \begin{pmatrix}
        \bm{\xi}_{\theta}^{\wedge} & \bm{\xi}_{v} & \bm{\xi}_p\\
        \bm{0}_{1,3} & 0 & 0 \\
        \bm{0}_{1,3} & 0 & 0       
    \end{pmatrix}
    \label{eq:se_2(3)wedge}
\end{equation}
\noindent where $\bm{\xi}_\theta^\wedge \in \mathfrak{so}(3)$ is the vector $[\delta\xi_{\theta,x}, \delta\xi_{\theta,y}, \delta\xi_{\theta,z}]^\top$ mapped as an element of the set of skew-symmetric $3\times3$ matrices that represent the angular velocity and also the rotation matrix $\bm{R}$ in its Lie algebra.
This group is useful to represent UAVs' states $\bm{R}$ (attitude), $\bm{v}$ (velocity), and $\bm{p}$ (position), and is used in \cite{Cohen_2020_lqr} to formulate a low-level quadcopter control system.

\subsection{Nomenclature}\label{sub-sec:nomenclature}
Consider the world frame, $\mathscr{F}^w$, as the inertial frame with orthonormal basis $\{\bm{x}_w,\bm{y}_w,\bm{z}_w\}$ in $\mathscr{F}^w$. The body frame, $\mathscr{F}^b$, is attached to the quadcopter's Center of Mass (CoM) and moves with the body. It also has an orthonormal basis $\{\bm{z}_b,\bm{y}_b,\bm{z}_b\}$ represented in $\mathscr{F}^w$. The Direction Cosine Matrix (DCM) is represented as $\bm{R}_w^b \in$ SO(3). Body position and velocity, expressed in the world frame, are represented as $\bm{p}_w$ and $\bm{v}_w$, and in the body frame as $\bm{p}_b$ and $\bm{v}_b$\footnote{Note that the subscript represents the frame in which the respective vector is represented}. The relation between the two is $\bm{p}_w=\bm{R}_w^b\bm{p}_b$. The angular velocity of the body rotating about the fixed frame, represented in the body frame, is denoted as $\bm{\omega}_b$.

Here, we also define the reference frame $\mathscr{F}^r$ as a frame that translates and rotates along with the reference trajectory of the quadcopter, given by a reference position $\bm{p}_{w}^{r}$, velocity $\bm{v}_{w}^{r}$, and acceleration $\bm{a_{w}}^{r}$. The matrix $\bm{R}_w^r$ represents the orientation of $\mathscr{F}^r$ with respect to $\mathscr{F}^w$. The angular velocity of the reference rotating about the fixed frame, represented in the reference frame, is denoted as $\bm{\omega}_r$.

\subsection{Dynamics of the Quadcopter}\label{sub-sec:quadcopter}
The dynamic model used in this work is presented in \cite{faessler-quadrotor-2018} and considers that the quadcopter does not have rotor drag and wind effects. It also considers that the thrust does not depend on the rotor drag, and the propellers are stiff. The time derivatives of the position, the velocity, the orientation matrix, and the angular velocity are given as

\begin{equation}\label{eq:pos-derivative-quad}
    \dot{\bm{p}_w} = \bm{v}_w
\end{equation}
\begin{equation}\label{eq:vel-derivative-quad}
    \dot{\bm{v}_w} = -g\bm{z}_w + \frac{f^T\bm{z}_b}{m_q} - \frac{1}{m_q}\bm{R}_w^b\bm{Dv}_b
\end{equation}
\begin{equation}\label{eq:R-derivative-quad}
    \dot{\bm{R}_w^b}=\bm{R}_w^b\bm{\omega}_b^{\wedge}
\end{equation}
\begin{equation}\label{eq:omega-derivative-quad}
    \dot{\bm{\omega}_b}=\bm{J}_b^{q^{-1}}(\bm{\tau}_b - \bm{\omega}_b \times\bm{J\omega}_b^{q^{-1}}-\bm{E}\bm{v}_b - \bm{F\omega}_b),
\end{equation}

\noindent where $m_q$ is the quadrotor's mass, $f^T$ is the thrust force, $\bm{\tau}_b$ is the torque, $\bm{z}_b =[0\text{ }0\text{ }1]^\top$, in the body frame,  $\bm{\omega}_b^{\wedge}$ is the skew-symmetrical matrix of $\bm{\omega}_b$ (similar to what is done in \eqref{eq:se_2(3)wedge}), $\bm{J}_b$ is the quadrotor's second moment of mass, $g=9.81$ m/s$^2$ is the gravity. The matrix $\bm{D}$ is the diagonal matrix containing the rotor-drag coefficients $d_x$ ,$d_y$, and $d_z$. The matrices $\bm{E}$ and $\bm{F}$ represent the drag in the rotational dynamics.

\section{Controller in the Special Euclidean Group}\label{sec:controller-model}

\begin{figure*}
    \centering
    \includegraphics[width=\textwidth]{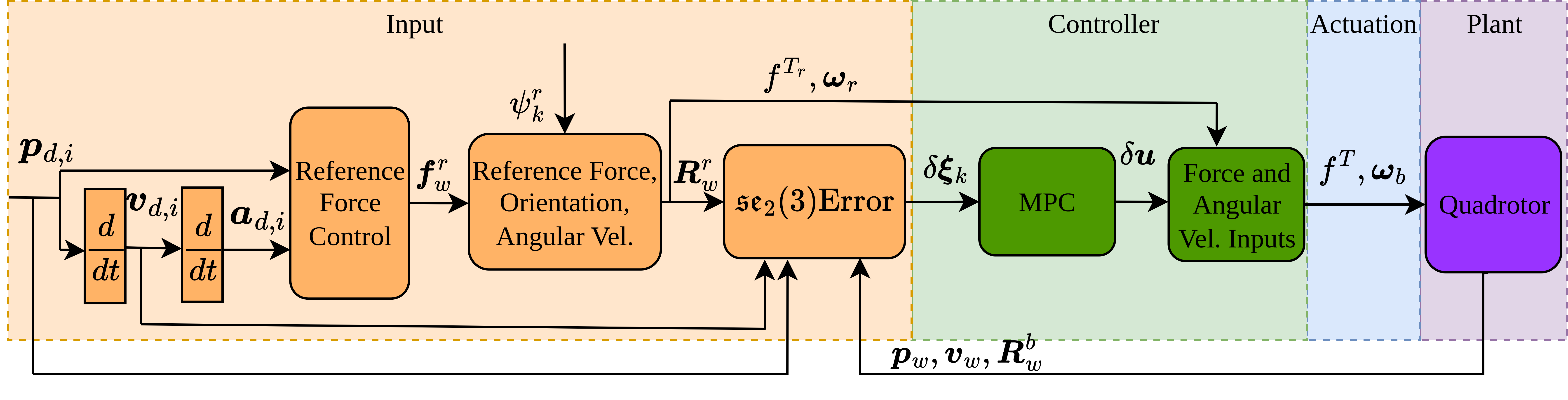}
    \caption{Block diagram containing the control system used in this work to control a real quadcopter}
    \label{fig:controller-diagram}
\end{figure*}
This section introduces how the work in \cite{Cohen_2020_lqr} models the error of the quadcopter states, and the state-space representation of the error dynamics in the Lie group SE$_2$(3). The original work used this model to solve an LQR problem, and our work uses it to solve an MPC problem.

Given a reference trajectory for a quadcopter, composed of position $\bm{p}_{w}^{r}$, velocity $\bm{v}_{w}^{r}$, and attitude $\bm{R}_w^r$. In SE$_2$(3), this trajectory is represented as
\begin{equation}
        \bm{X}^r = \begin{bmatrix}
        \bm{R}_w^r & \bm{v}_{w}^{r} & \bm{p}_{w}^{r}\\
        \bm{0} & 1 & 0 \\
        \bm{0} & 0 & 1
    \end{bmatrix}.
\end{equation}

A low-level trajectory tracking SE$_2$(3) LQR controller~\cite{Cohen_2020_lqr} uses the left-invariant tracking error calculated as
\begin{equation}\label{eq:inv-error}
    \delta\bm{X} = \bm{X}^{-1}\bm{X}^r = \begin{bmatrix}
        \delta\bm{R} & \delta\bm{v} & \delta\bm{p}\\
        \bm{0} & 1 & 0 \\
        \bm{0} & 0 & 1
    \end{bmatrix}
\end{equation}
where $\bm{X}$ are the quadrotor states in SE$_2$(3) and
\begin{eqnarray}
    \delta\bm{R} &=& \bm{R}_w^{r^\top}\bm{R}_w^b, \label{eq:error-R}\\
    \delta\bm{v}&=&\bm{R}_w^{b^\top}(\bm{v}_{w}^{r} - \bm{v_{w}}),\label{eq:error-v}\\
    \delta\bm{p} &=& \bm{R}_w^{b^\top}(\bm{p}_{w}^{r} - \bm{p_{w}}).\label{eq:error-p}
\end{eqnarray}

The inputs to the system are
\begin{equation}
    \bm{u} = [f_b^T \text{ } \bm{\omega}_b]^\top,
\end{equation}
where $f_b^T$ is the thrust, and $\bm{\omega}_b$ is the angular velocity in the body frame.


Given a reference trajectory ($\bm{p}_{w}^{r}$, $\psi_{w}(t)$), where $\psi_{w}(t)$ is the reference angle of yaw, to compute the system inputs ($\bm{u}$) one may follow the process of:

\begin{enumerate}[label=\Roman*.]
    \item compute $\bm{v}_{w}^{r}$, $\bm{R}_w^r$, $f^{T_r}$, and $\bm{\omega}_r$ (reference velocity, attitude, thrust, and angular velocity) following~\cite{Cohen_2020_lqr};

    \item compute the left-invariant error in \eqref{eq:inv-error} transforming it into an element of $\mathfrak{se}_2$(3) using 
\begin{equation}
   \delta\bm{\xi}^{\wedge}=\log(\delta\bm{X}) = [\delta\bm{\xi}_{\phi} \text{ } \delta\bm{\xi}_v \text{ }\delta\bm{\xi}_p]^\wedge .
   \label{eq:algebra-error}
\end{equation}
\end{enumerate}


Finally, the continuous-time error dynamics are written as
\begin{equation}\label{eq:error-se23}
    \delta\dot{\bm{\xi}} = \bm{A}\delta\bm{\xi} + \bm{B}\delta\bm{u},
\end{equation}
where
\begin{equation}\label{eq:inputs-quad}
    \delta\bm{u}=\begin{bmatrix}
        \delta f^T \\ \delta\bm{\omega}_{b}
    \end{bmatrix}=
    \begin{bmatrix}
        f^{T_r} - f^T\\
        \delta\bm{R}\bm{\omega}_{r} - \bm{\omega}_{b}
    \end{bmatrix}.
\end{equation}
The matrices $\bm{A} \in \mathbb{R}^{n\times n}$ and $\bm{B}\in \mathbb{R}^{n\times m}$, where $n$ is the number of states and $m$ is the number of inputs, are obtained with a linearization about the desired trajectory~\cite{Cohen_2020_lqr}. This involves differentiating \eqref{eq:error-R}, \eqref{eq:error-v}, and \eqref{eq:error-p} with respect to time, combining the results with \eqref{eq:vel-derivative-quad}, and applying Lie group properties under the assumption of small $\delta\bm{\xi}$. Since the linearized matrices representation ends up being a function of the desired trajectory only, the matrices can be computed offline and are discretized using Euler's method.

\cite{Cohen_2020_lqr} also augment the state space with an integral term to reduce steady-state errors from linearization and parameter uncertainties. This additional state is defined as
\begin{equation}
    \bm{\xi}^{int}=\int_0^t(c_1\delta\bm{p} + \delta\bm{v})d\tau
\end{equation}
where $c_1>0$ is a constant. In this case, $\delta\bm{\xi}^{aug}= [\delta\bm{\xi}^{\vee}_{\phi} \text{ } \delta\bm{\xi}_v \text{ }\delta\bm{\xi}_p \text{ }\delta\bm{\xi}^{int}]^\top$. The linearized dynamics is

\begin{equation}\label{eq:error-state-aug}
\underbrace{
\begin{bmatrix}
\delta \boldsymbol{\xi}^{\phi} \\
\delta \boldsymbol{\xi}^{v} \\
\delta \boldsymbol{\xi}^{r} \\
\delta \boldsymbol{\xi}^{int}
\end{bmatrix}
}_{\delta \boldsymbol{\xi}^{aug}}
=
\underbrace{
\begin{bmatrix}
0 & 0 & 0 & 0 \\
\mathbf{A}_{2,1} & \mathbf{A}_{2,2} & 0 & 0 \\
0 & 1 & -\bm{\omega}_r^\wedge & 0 \\
0 & 1 & c_1 \bm{I} & 0
\end{bmatrix}
}_{\mathbf{A}_{aug}}
\underbrace{
\begin{bmatrix}
\delta \boldsymbol{\xi}^{\phi} \\
\delta \boldsymbol{\xi}^{v} \\
\delta \boldsymbol{\xi}^{r} \\
\delta \boldsymbol{\xi}^{int}
\end{bmatrix}
}_{\delta \boldsymbol{\xi}^{aug}}
+
\underbrace{
\begin{bmatrix}
\mathbf{B} \\
0
\end{bmatrix}
}_{\mathbf{B}_{aug}}
\delta \mathbf{u},
\end{equation}

\noindent where the elements $\mathbf{A}_{2,1}$ and $\mathbf{A}_{2,2}$ are given by
\begin{align}\label{eq:low-level-A-elements}
\mathbf{A}_{2,1} &= \frac{1}{m_q} \big( \left( \mathbf{D} \mathbf{C}_{ar}^\top \mathbf{v}_a^{z_r w/a} \right)^\wedge \\ \nonumber
 & - \mathbf{D} \left( \left( \mathbf{C}_{ar}^\top \mathbf{v}_a^{z_r w/a} \right)^\wedge - \left( f_r^\top \mathbf{z}_{b} \right)^\wedge \right) \big),  \\
\mathbf{A}_{2,2} &= -\bm{\omega_r}^\wedge - \frac{\mathbf{D}}{m_q},
\end{align}
and 
\begin{equation}
\bm{B}=
    \begin{bmatrix}
        \bm{0}&\bm{I}\\
        \frac{1}{m_q}\bm{z}_b&\bm{0}\\
        \bm{0}&\bm{0}
    \end{bmatrix}
\end{equation}

For the system without the augmented state, $n=9$ and $m=3$. With the augmented state, $n=12$.

\subsection{Model Predictive Control}\label{sub-sec:mpc}

The error MPC problem used here is stated as
\begin{equation}\label{eq:mpc-cost}
    \min_{\delta \bm{u}_0,...\delta \bm{u}_{N_h}} \sum_{k=1}^{N_h} \left(\delta\bm{\xi}_k^\top \bm{Q}\delta\bm{\xi}_k + \delta\bm{u}_k^\top\bm{R}\delta \bm{u}_k\right)   
\end{equation}
subject to
\begin{align}\label{eq:mpc-restriction-1}
\bm{\xi}_{k+1} = \bm{A}_k\bm{\xi}_k + \bm{B}_k\delta\bm{u}_k\\    
\delta\bm{\xi}_0 = \log{(\delta\bm{X}_0)}\\
\delta\bm{u}_{min}\leq \delta\bm{u}_k \leq  \delta\bm{u}_{max}
\end{align}
where $\bm{Q}$ and $\bm{R}$ are error and input weight matrices, $\delta\bm{\xi}_0$ is the initial left invariant error, $\delta\bm{u}_{min}$ and $\delta\bm{u}_{max}$ are the constraints on the input error, and $N_h$ is the prediction horizon. 

Equation \eqref{eq:mpc-cost} can be rearranged using \eqref{eq:error-se23} to represent the MPC problem in matrix form as
\begin{equation}
    \min_{\Delta\bm{U}} \frac{1}{2}\Delta\bm{U}^\top\bm{H}\Delta\bm{U} + \Delta\bm{U}^\top\bm{G}
\label{eq:mpc-matrix-rep}
\end{equation}
subject to
\begin{equation}
    \Delta\bm{U}_{min} \leq \Delta\bm{U} \leq \Delta\bm{U}_{max}   
\end{equation}
where 
\begin{equation}
    \bm{H} = 2(\bm{\mathcal{B}}^\top\bar{\bm{Q}}\bm{\mathcal{B}}+\bar{\bm{R}}),
\end{equation}
\begin{equation}
    \bm{G} = 2\bm{\mathcal{B}}^\top\bar{\bm{Q}}\bm{\mathcal{A}}^\top\delta\bm{\Xi}_k,
\end{equation}
\begin{equation}
    \bm{\mathcal{A}} = \begin{bmatrix}
        \bm{A}&
        \bm{A}^2&
        \dots&
        \bm{A}^N
    \end{bmatrix}^\top
    \in \mathbb{R}^{N_hn\times n},
\end{equation}
\begin{equation}
    \bm{\mathcal{B}} = 
\begin{bmatrix}
\bm{B} & \bm{0} & \bm{0} & \cdots & \bm{0} \\
\bm{AB} & \bm{B} & \bm{0} & \cdots & \bm{0} \\
\bm{A}^2\bm{B} & \bm{AB} & \bm{B} & \cdots & \bm{0} \\
\vdots & \vdots & \vdots & \ddots & \vdots \\
\bm{A}^{N-1}\bm{B} & \bm{A}^{N-2}\bm{B} & \bm{A}^{N-3}\bm{B} & \cdots & \bm{B}
\end{bmatrix},
\end{equation}
\begin{equation}\label{eq:Q-bar-mpc}
    \bar{\bm{Q}} = diag(\bm{Q},\bm{Q},...,\bm{Q})\in \mathbb{R}^{N_hm\times N_hm}
\end{equation}
\begin{equation}\label{eq:R-bar-mpc}
    \bm{\bar{R}} = diag(\bm{R},\bm{R},...,\bm{R}) \in \mathbb{R}^{N_hn\times N_hm}.
\end{equation}
Additionally, $\delta\bm{\Xi}_k\in \mathbb{R}^{N_hn}$, $\Delta\bm{U}\in \mathbb{R}^{N_hm}$, and $\Delta\bm{U}_{min},\Delta\bm{U}_{max}\in \mathbb{R}^{N_hm}$ are the stacked matrices of error, input error, and input constraints. The quadratic program in~\eqref{eq:mpc-matrix-rep} is solved to find the inputs $\Delta\bm{U}$ that minimize the cost over time horizon $N_h$. The first $m$ elements of the stacked matrix $\Delta\bm{U}$ are used as input to the quadcopter.

\subsection{System-Level Architecture}

Fig.~\ref{fig:controller-diagram} provides an overview of the proposed control architecture. The input of the system (orange boxes) takes the reference trajectory ($\bm{p}_{w}^{r}$, $\psi_{w}(t)$) to compute the quadcopter's reference states and inputs. It also uses the states to calculate the $\mathfrak{se}_2$(3) error $\delta\bm{\xi}$ according to \eqref{eq:error-se23} and \eqref{eq:algebra-error}. 

This error feeds into the controller (green area), which outputs $\delta\bm{u}$ -- the correction needed to minimize $\delta\bm{\xi}$. The controller can be any linear controller, such as PID, LQR, or MPC. We evaluate the performance of these approaches through simulation and real-world experiments.

The controller's actuations (blue area) are calculated isolating $f^T$ and $\bm{\omega}_b$ in \eqref{eq:inputs-quad} and sent to the plant (purple area). While \cite{Cohen_2020_lqr} used a simulated plant, our work uses a real quadcopter and analyzes how the behavior of the system changes from the simulated version.

\section{Experiments}\label{sec:experiments}

This section presents simulation and real-world experiments to validate the proposed architecture. 
We show: \textit{i)} simulated results comparing our SE$_2$(3) MPC to the baseline LQR~\cite{Cohen_2020_lqr}, and 
\textit{ii)} real flight tests evaluating the \textit{baseline}, our approach, and the industry-standard controller. 
In both cases, the reference trajectory  was a $1$~m radius circle with a $30$~s  period.

\subsection{Simulation Setup and Parameters}\label{sec:simulation}

Two simulation trials were performed. \textit{Sim-I} tested the \textit{baseline} approach and \textit{Sim-II} evaluated the performance of our approach, the SE$_2$(3) MPC controller. Both experiments were performed on a simulated quadrotor platform, using \eqref{eq:pos-derivative-quad}, \eqref{eq:vel-derivative-quad}, \eqref{eq:R-derivative-quad}, and \eqref{eq:omega-derivative-quad}. The model parameters, such as the drag matrices $\bm{D}, \bm{E}$, and $\bm{F}$, as well as the mass $m_q$ and $\bm{J}_b$ were the same as presented in \cite{Cohen_2020_lqr}, for fair comparison with the original performance of the \textit{baseline} controller.

Model mismatch was introduced in the simulation to test the effectiveness of the augmented state and for a better comparison with the results obtained in the real vehicle. The drag matrix $\bm{D}$ was considered null when calculating the matrices $\bm{A}_{aug}$ and $\bm{B}_{aug}$, but drag was added to the dynamics. 

For the \textit{baseline} approach, the LQR weight matrices were $\bm{Q}_\theta= 10\cdot\bm{I}$, $\bm{Q}_v = 10\cdot\bm{I}$, $\bm{Q}_p = 100\cdot\bm{I}$, $\bm{Q}_{aug} = 0.1\cdot\bm{I}$, $\bm{Q} = \operatorname{diag}(\bm{Q}_\theta,\bm{Q}_v,\bm{Q}_p,\bm{Q}_{aug})$, and $\bm{R} = \operatorname{diag}(Q_f,Q_\omega,Q_\omega,Q_\omega)$, where $Q_f = 0.05$, and $Q_\omega = 2.5$. The augmented state parameter, $c_1$, was chosen as $0.01$. The simulation time step was dt=$0.002$ s.

The prediction horizon used for the SE$_2$(3) MPC was $N_h=10$. The state error weights were $\bm{Q}_\theta = 10^5\cdot\bm{I}$, $\bm{Q}_v = 5*10^6\cdot\bm{I}$, $\bm{Q}_p = 10^9\cdot\operatorname{diag}(7,9,4)$. The MPC weights matrix was $\bar{\bm{Q}} = \operatorname{diag}(\bm{Q}_\theta,\bm{Q}_v,\bm{Q}_p)$. The input error weights matrix was $\bar{\bm{R}} = \operatorname{diag}(R_f,R_\omega,R_\omega,R_\omega)$, where $R_f = 10^3$ and $R_\omega=10$. The augmented state was not used to speed up the computation, as it was slowed by the quadratic optimization that the MPC performs at each time step. Finally, the simulation time step was dt=$0.002$ s. 
The parameters of both controllers were carried over to the real-world experiments without change unless otherwise discussed.

\subsection{Real Experiment Parameters}\label{sec:real-experiments}
\begin{figure}
    \centering
    \subfloat[Test area\label{fig:test-area}]{\includegraphics[width=0.4\linewidth]{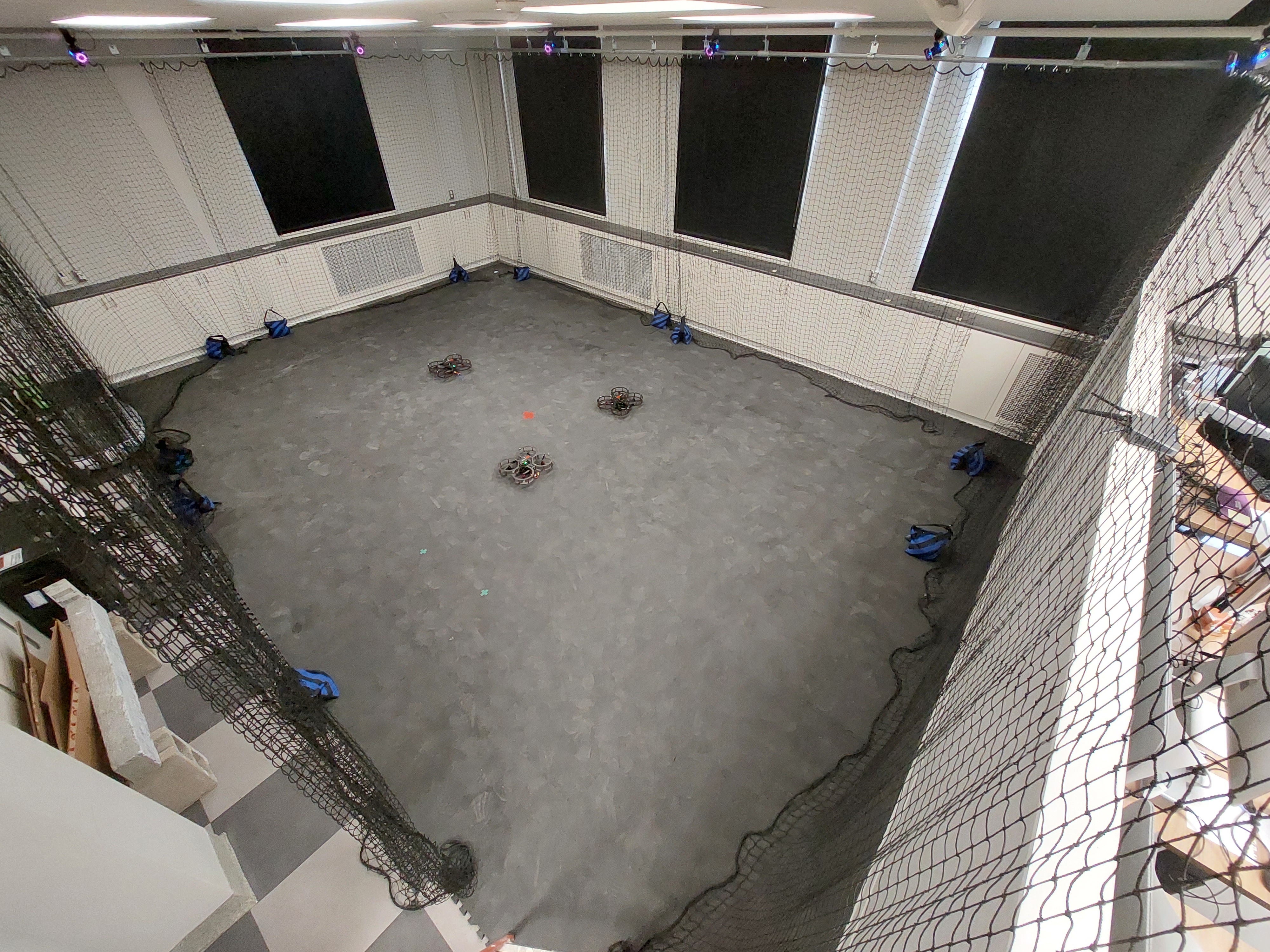}}
    \quad
    \subfloat[Quanser QDrone\label{fig:qdrone}]{\includegraphics[width=0.5\columnwidth]{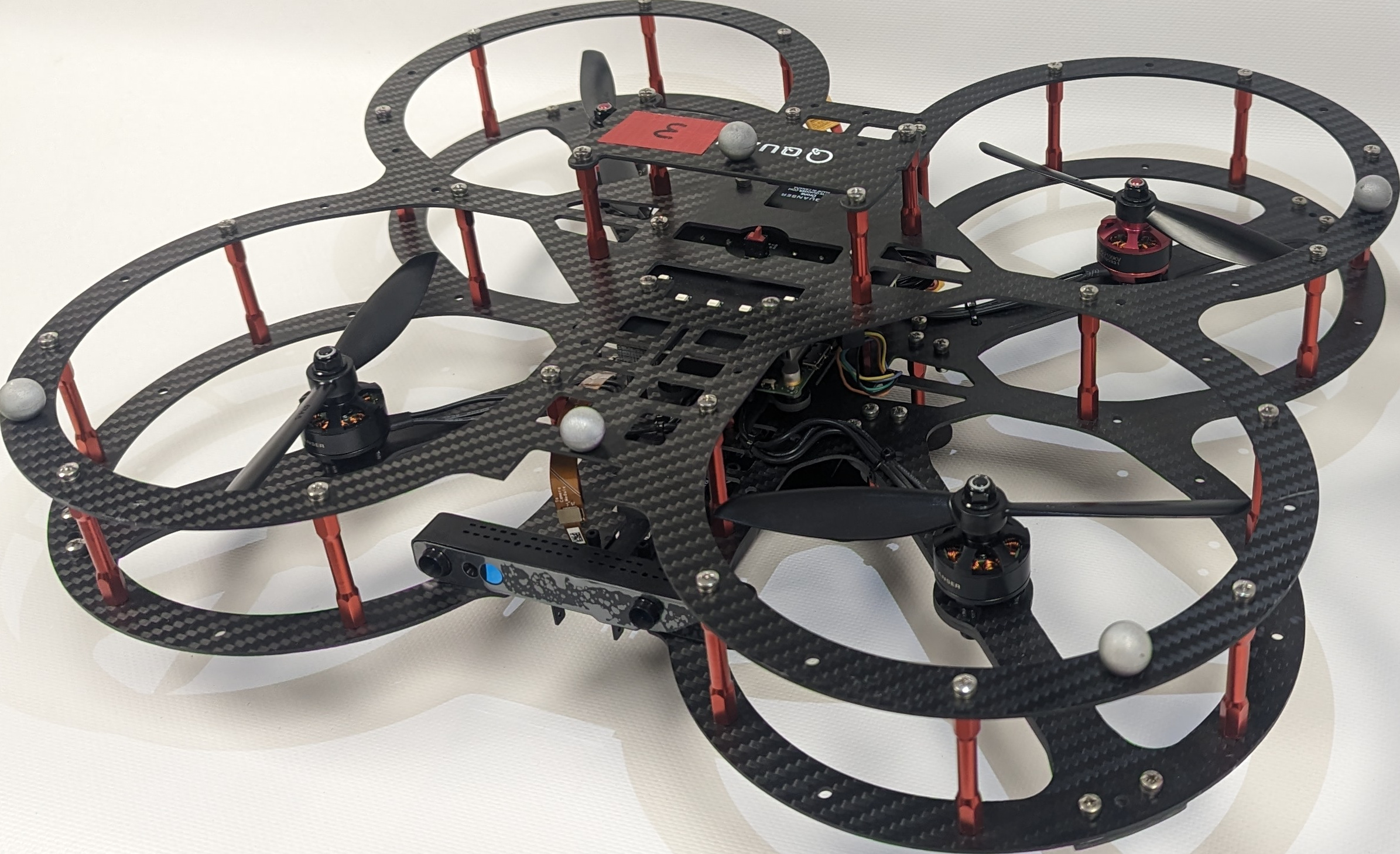}}
    \caption{Test area used for all the experiments in the real UAV in this work, and the quadcopter used for the experiments}
    
\end{figure}

The UAV flew in an indoor environment of dimensions $5 \times 7.5 \times 3$~m (width, length, and height, respectively), shown in Fig.~\ref{fig:test-area}. The quadcopter used was the QDrone (Fig.~\ref{fig:qdrone}), provided by Quanser.

UAV position and orientation were tracked using a $10$-camera Vicon System\footnote{\url{www.vicon.com}}, which computed the states at $100$~Hz and transmitted them to the vehicle via a ground station. Velocity was estimated using the UAV's internal sensors.

The test environment uses an inertial coordinate system with the $X$ and $Y$ axes defining the horizontal plane and $Z$ pointing upward. Coordinates $x$, $y$, $z$ correspond to these axes, with the origin at ground level at the center of the room.

Due to the unknown drag matrix of the QDrone and minimal wind in the test area, drag was neglected in \eqref{eq:low-level-A-elements} when computing $\bm{A}_{aug}$.

Three experiments were conducted. \textit{Exp-I} used the QDrone's default controller --  two cascaded proportional integral velocity (PIV) controllers -- where the first stage maps position error and current velocity to desired thrust and attitude, and the second stage takes attitude error as input and outputs desired angular velocities. \textit{Exp-II} validated the \textit{baseline} SE$_2$(3) LQR~\cite{Cohen_2020_lqr} on real vehicles, which was not done in the original strudy. \textit{Exp-III} validated the SE$_2$(3) MPC proposed in this work. All experiments used QDrone's default low-level motor controllers to ensure stable flight and fair comparison.

\section{Results} \label{sec:results}



\subsection{Simulation}

\begin{figure}
    \centering
    \subfloat[SE$_2$(3) LQR\label{fig:3D-lqr-sim}]{    \includegraphics[width=0.5\linewidth]{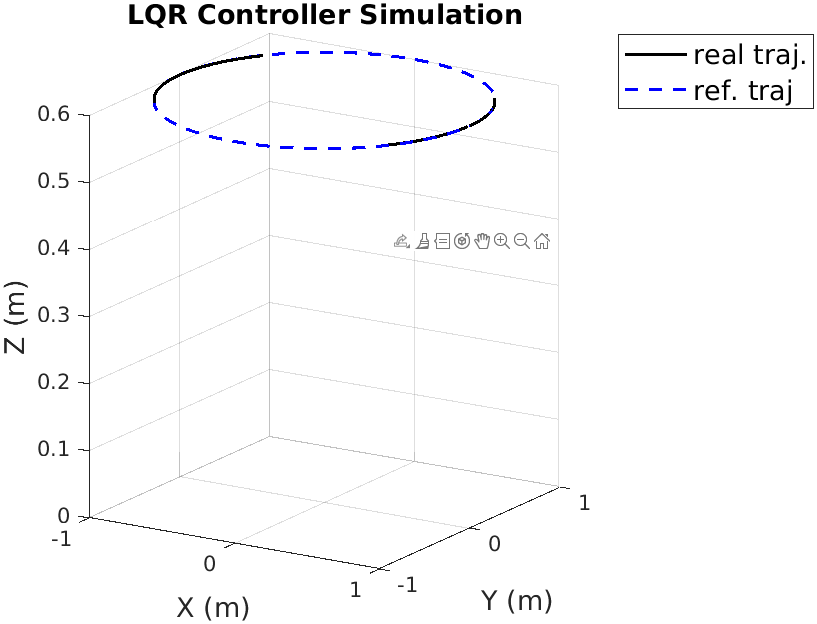}}
    \subfloat[SE$_2$(3) MPC\label{fig:3D-mpc-sim}]{\includegraphics[width=0.5\linewidth]{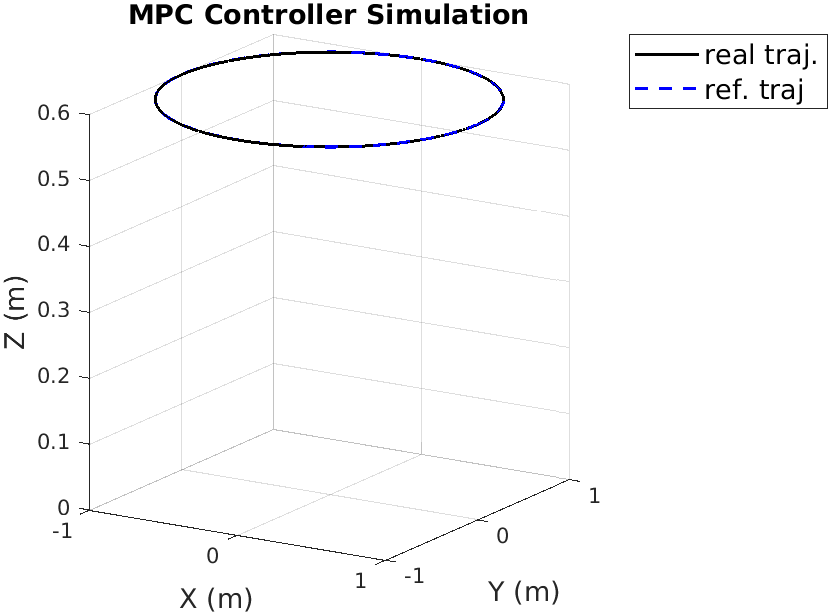}}
    \caption{(a) Simulated behavior of the \textit{baseline} SE$_2$(3) LQR vs. (b) simulated behavior of the proposed SE$_2$(3) MPC.}   
\end{figure}

\textit{Sim-I}: Fig~\ref{fig:3D-lqr-sim} shows the 3D reference and simulated trajectory using the SE$_2$(3) LQR. \textit{Sim-II}: Fig.~\ref{fig:3D-mpc-sim} displays the result for our SE$_2$(3) MPC.

Both controllers successfully tracked the circular path with smooth trajectories, demonstrating that  our approach performs comparably to a state-of-the art technique, validating it in simulation.

\subsection{Real-world Experimental Validation}

\begin{figure}
    \centering
    \subfloat[Default QDrone Controller\label{fig:3D-quanser}]{\includegraphics[width=0.5\linewidth]{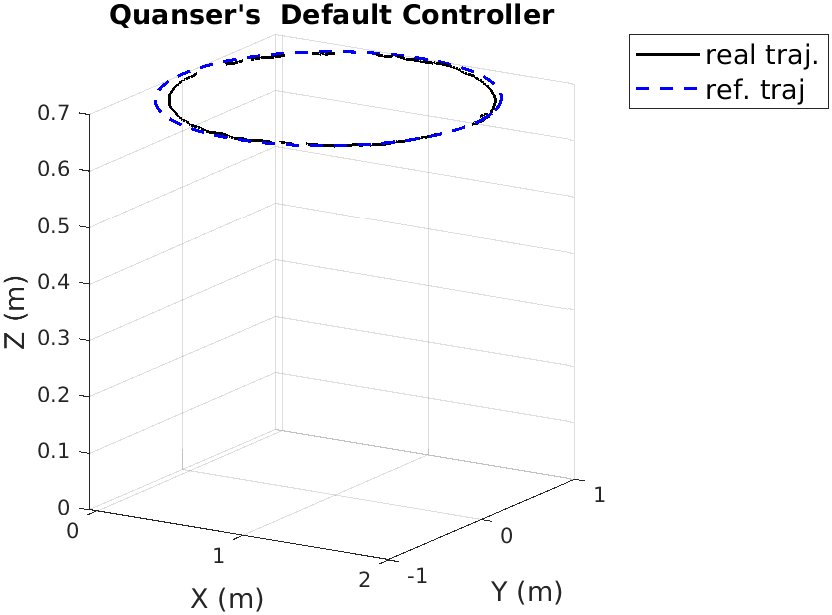}}
    \subfloat[\textit{Baseline} SE$_2$(3) LQR \label{fig:3D-lqr-exp}]{\includegraphics[width=0.5\linewidth]{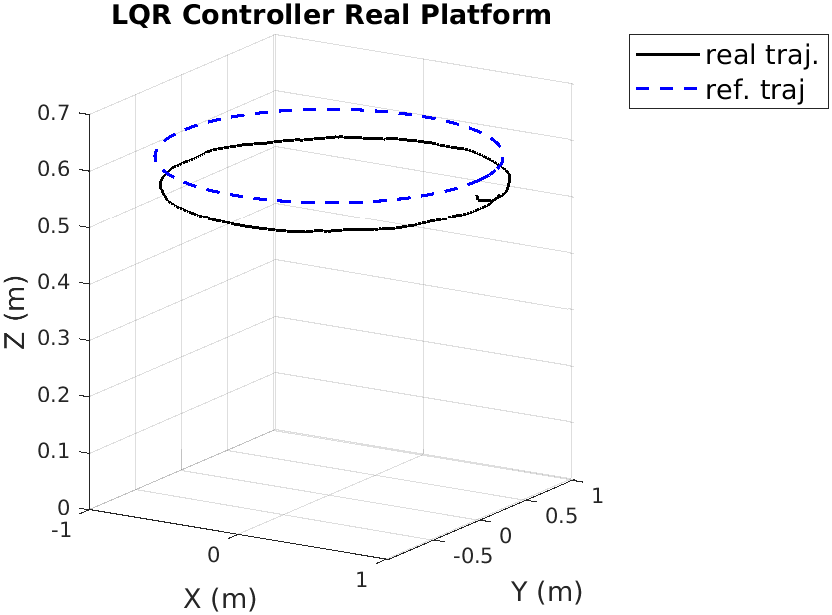}}
    \quad
    \subfloat[SE$_2$(3) MPC\label{fig:3D-mpc-exp}]{\includegraphics[width=0.5\linewidth]{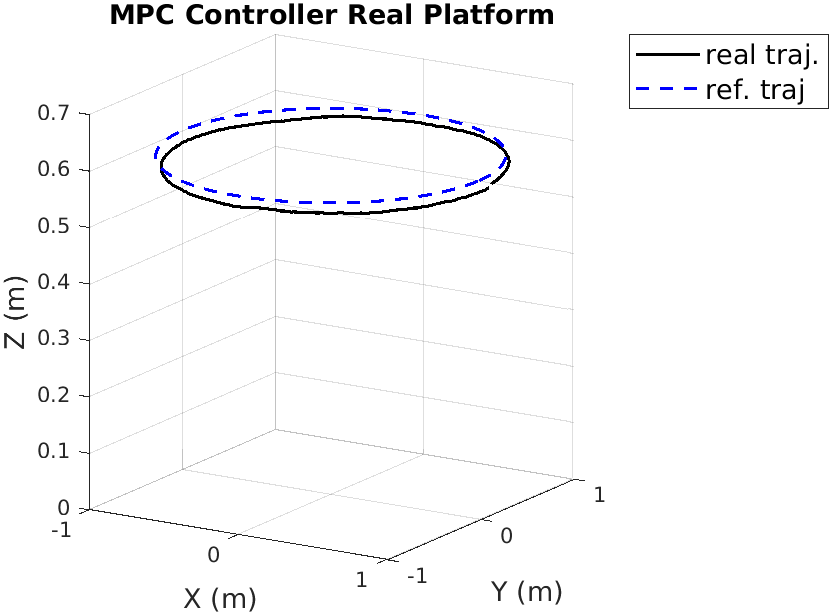}}
    \caption{QDrone 3D trajectory controlled by (a) Default QDrone Controller, (b) \textit{baseline} SE$_2$(3) LQR, and (c) SE$_2$(3) MPC.}

\end{figure}

\begin{figure}
    \centering
    \includegraphics[width=0.6\linewidth]{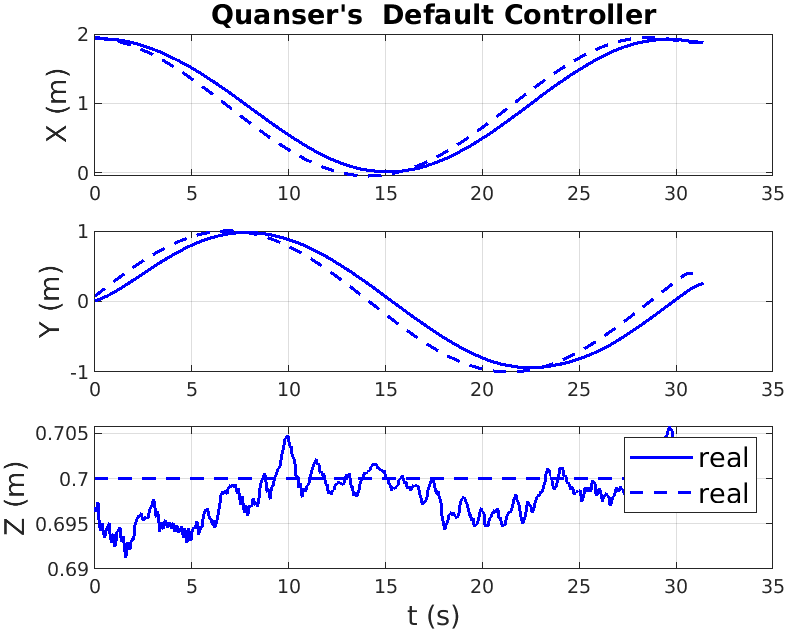}
    \caption{$X$, $Y$, and $Z$ trajectories over time, executed by Qdrone when controlled by the Default QDrone Controller.}
    \label{fig:xyz-quanser}
\end{figure}

\textit{Exp-I}: Fig.~\ref{fig:3D-quanser} shows the 3D trajectory executed by the Default QDrone Controller, with corresponding  $x$, $y$, and $z$ in~Fig.~\ref{fig:xyz-quanser}. The controller failed to follow the reference trajectory precisely in the $X$ and $Y$ axes, while $Z$ tracking was satisfactory, with low overshoot and steady state errors. -- though the $z$ coordinate exhibited noticeable noise, as will be discussed in later comparisons.

\textit{Exp-II}: Fig.~\ref{fig:3D-lqr-exp} shows the trajectory of the real quadcopter using the \textit{baseline} SE$_2$(3)LQR. Despite a small difference in the $z$ coordinate, the UAV closely followed the desired curve, confirming that the linearization in~\cite{Cohen_2020_lqr} effectively models the error dynamics of a real quadcopter.


\textit{Exp-III}: Fig.~\ref{fig:3D-mpc-exp} shows the real quadcopter 3D trajectory under our SE$_2$(3) MPC. Performance in $x$ and $y$ was similar to that of the \textit{baseline} LQR, but tracking in $z$ was notably more accurate.


\subsection{From Simulation to Real World Analysis}

\begin{figure}
    \centering
    \subfloat[Positions in the Simulation\label{fig:xyz-lqr-sim}]{\includegraphics[width=0.5\linewidth]{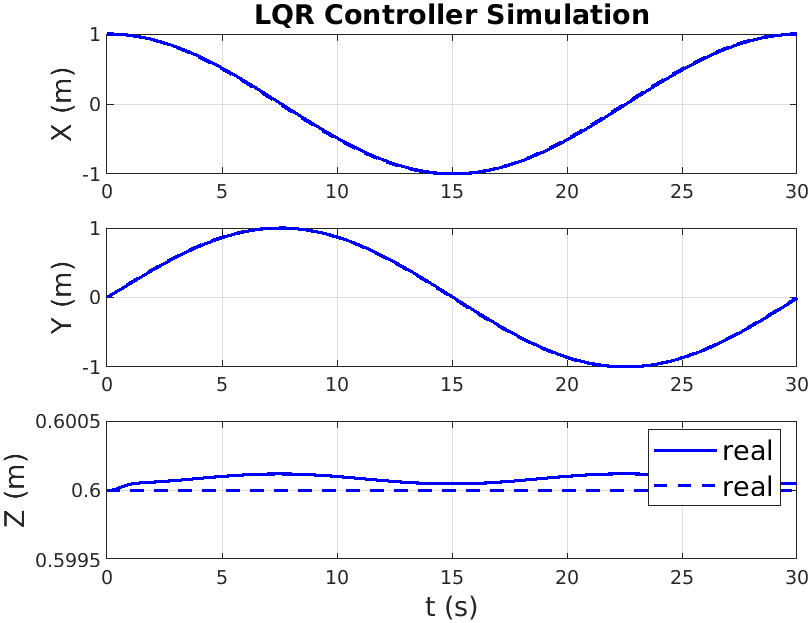}}
    \subfloat[Positions in the Real Quadcopter (QDrone)\label{fig:xyz-lqr-exp}]{    \includegraphics[width=0.5\linewidth]{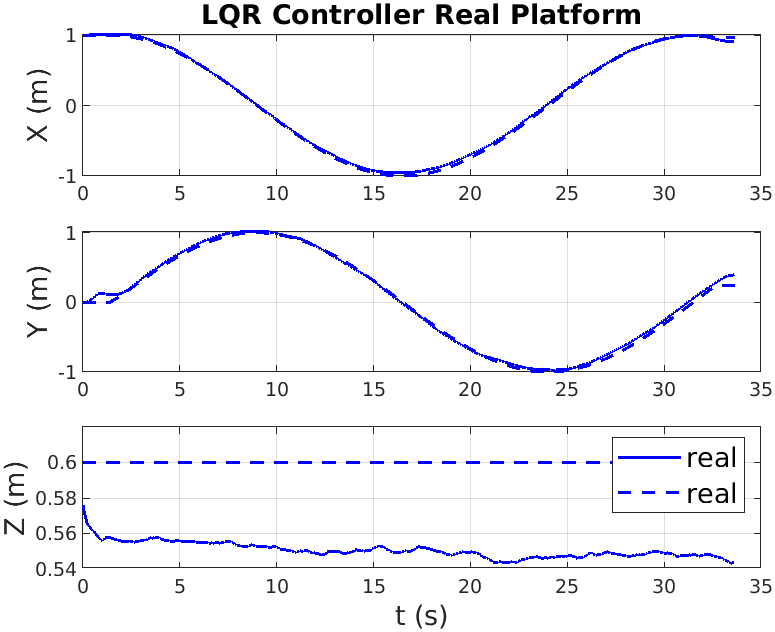}}
    \quad
    \subfloat[Errors in Simulation\label{fig:errors-lqr-sim}]{\includegraphics[width=0.5\linewidth]{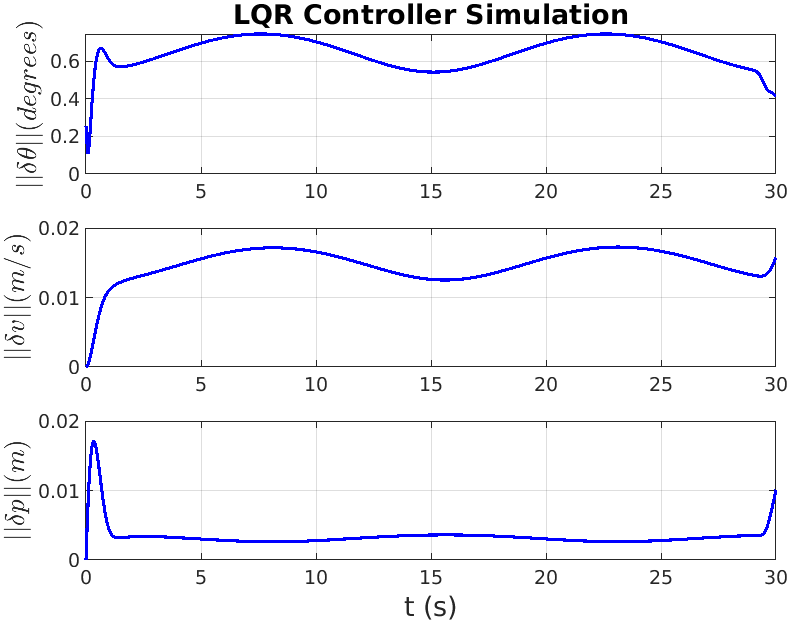}}
    \subfloat[Errors in the Real Quadcopter (QDrone)\label{fig:errors-lqr-exp}]{\includegraphics[width=0.5\linewidth]{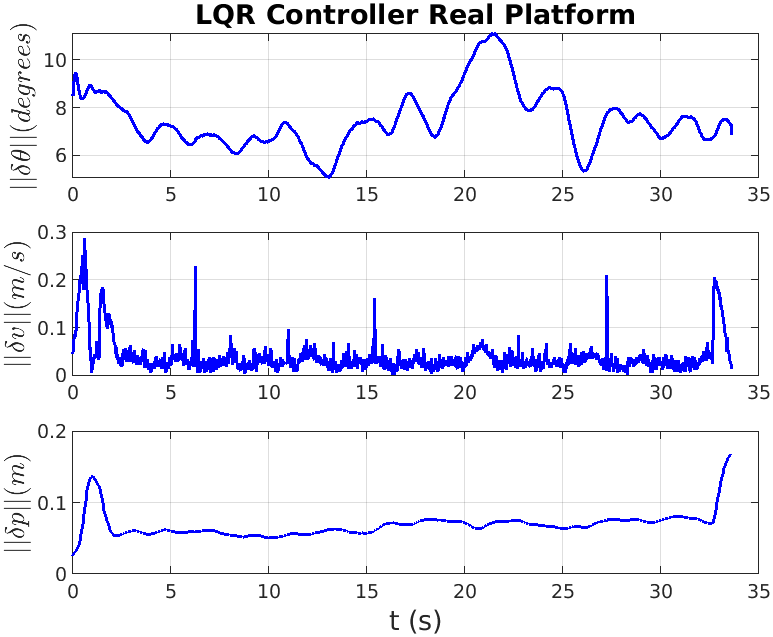}}
    \caption{Simulation and real-world results of the \textit{baseline} SE$_2$(3) LQR controller over time. (c) and (d) display position ($\|\delta\bm{p}\|$), velocity ($\|\delta\bm{v}\|$) and orientation ($\|\delta\bm{\theta}\|$) errors norms.}
\end{figure}

\textbf{Baseline SE$_2$(3) LQR Controller}:
Figs.~\ref{fig:xyz-lqr-sim} and~\ref{fig:xyz-lqr-exp} show the $x$, $y$, and $z$ coordinates in simulation and in the real quadcopter. A $5$~cm error in $z$ was observed due to the sim-to-real differences like noise and unmodelled dynamics. While better tuning could reduce this, the QDrone requires takeoff using its default low-level controller, and switching controllers during flight caused instability with larger $\bm{Q}_p$ and $\bm{Q}_{aug}$. Despite this, $x$ and $y$ tracking closely matched simulation performance. 


Figs.~\ref{fig:errors-lqr-sim} and~\ref{fig:errors-lqr-exp} show state errors in the simulation and in the real quadcopter. In simulation, the orientation error showed a small overshoot, stabilizing near $0.5^\circ$. The real quadcopter had no overshoot, but exhibited more noise compared to the simulation, ranging from $6^\circ$ to $10^\circ$. The velocity error did not present any overshoot and remained between $0.01$-$0.02$~m/s in simulation, while the quadcopter presented an overshoot and stabilized below $0.1$~m/s. Position error curves are similar in simulation and in the real quadcopter, with an overshoot at the beginning and steady-state error. However, the position error in the quadcopter was about $10\times$ larger than that in the simulation. 

These differences stem from real-world noise and unmodelled dynamics. Still, real position and velocity errors stayed within $10\%$ of the reference, respectively, validating the method's effectiveness.

The negligible impact of neglecting drag further supports the robustness of the controllers, as the simulation showed minimal position error despite this assumption (Fig.~\ref{fig:errors-lqr-sim}).

\begin{figure}
    \centering
    \subfloat[Positions in the Simulation\label{fig:xyz-mpc-sim}]{\includegraphics[width=0.5\linewidth]{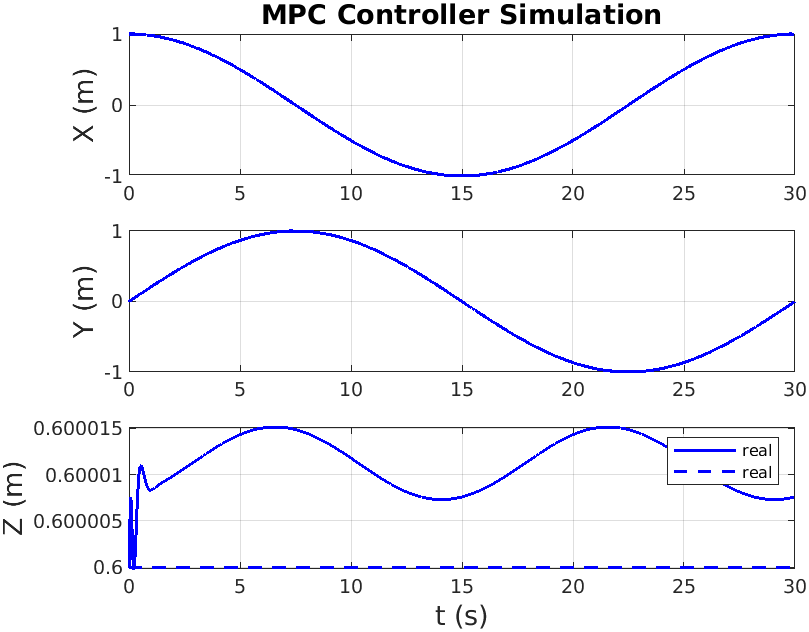}}
    \subfloat[Positions in the Real Quadcopter (QDrone)\label{fig:xyz-mpc-exp}]{    \includegraphics[width=0.5\linewidth]{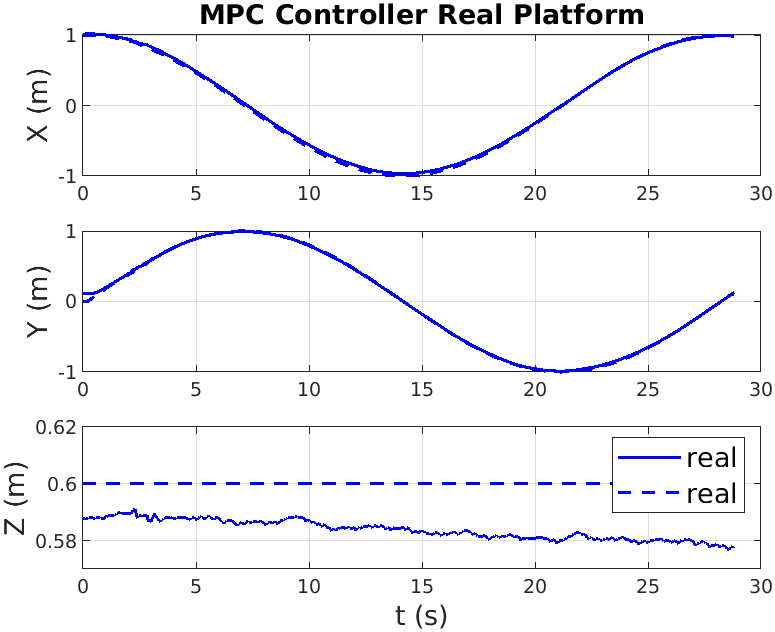}}
    \quad
    \subfloat[Errors in Simulation\label{fig:errors-mpc-sim}]{\includegraphics[width=0.5\linewidth]{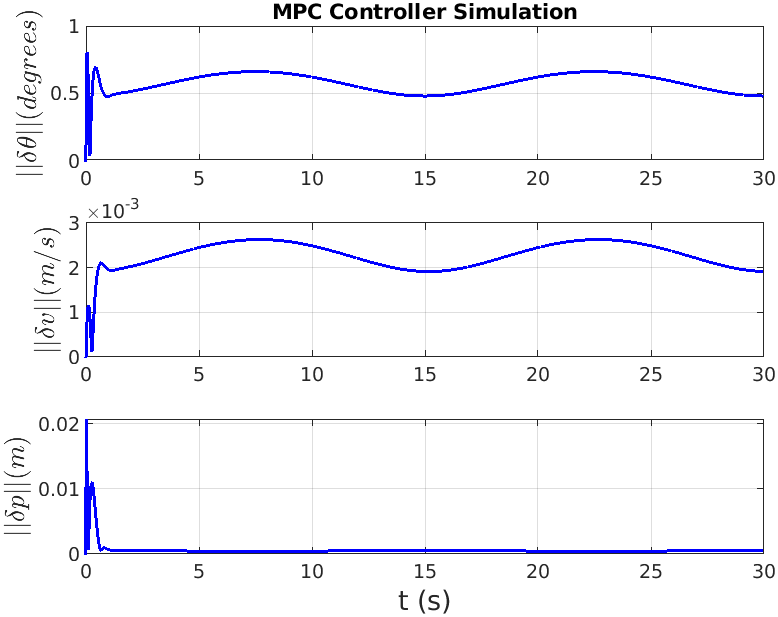}}
    \subfloat[Errors in the Real Quadcopter (QDrone)\label{fig:errors-mpc-exp}]{\includegraphics[width=0.5\linewidth]{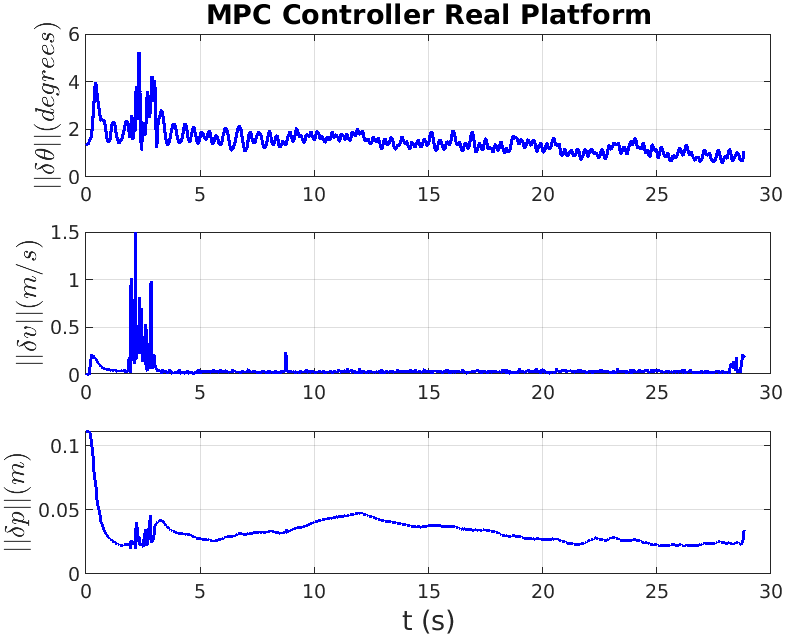}}
    \caption{Simulation and real-world results of the SE$_2$(3) MPC controller over time. (c) and (d) display position ($\|\delta\bm{p}\|$), velocity ($\|\delta\bm{v}\|$) and orientation ($\|\delta\bm{\theta}\|$) errors norms.}
\end{figure}

\textbf{SE$_2$(3) MPC controller}:
The $x$ and $y$ coordinates of the real quadcopter (Fig.~\ref{fig:xyz-mpc-exp}) behaved similarly to those of the simulation (Fig.~\ref{fig:xyz-mpc-sim}), though $z$ performance was slightly worse in the real case. 

Figs.~\ref{fig:errors-mpc-sim} and~\ref{fig:errors-mpc-exp} display the norm of the state errors over time for the MPC in simulation and on the real quadcopter. In simulation, the velocity error stabilized around $5\times10^{-3}$~m/s with no overshoot, and the position error had a small overshoot, settling at approximately $0.05$~m. The orientation error stabilized at approximately $0.5^\circ$.

In the real quadcopter, velocity error presented a high overshoot of about $0.6$~m/s before stabilizing near $0$~m/s. Position error had a small overshoot ($\approx0.1$~m) and stabilized around $0.05$~m, corresponding to a $5\%$ trajectory error (circle of $1$~m radius) -- significantly better than the $20\%$ seen with the commercial controller. The orientation error initially overshot but stabilized around $2^\circ$.

\subsection{Lie Group-based Control Strategies vs. Industry-Standard Solution in the Real World.}

\begin{figure}
    \centering
    \includegraphics[width=\linewidth]{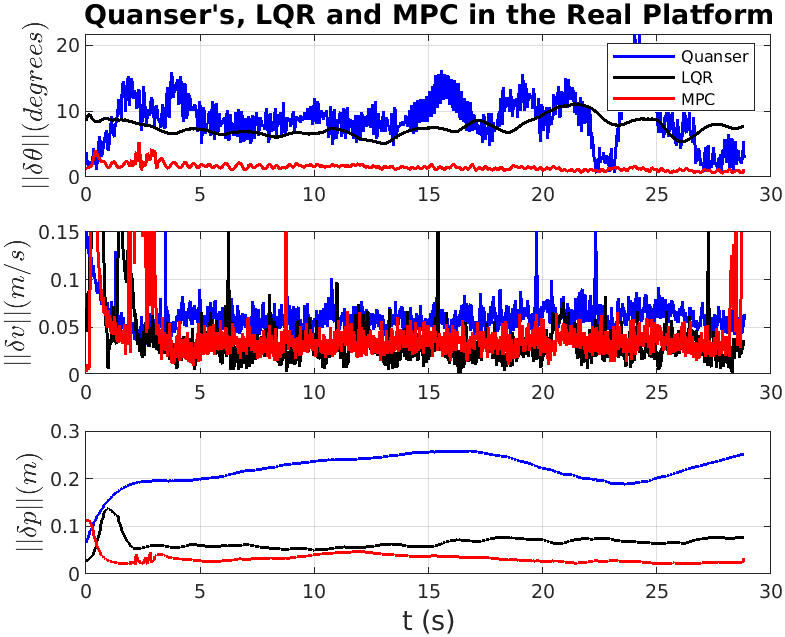}
    \caption{Position ($\|\delta\bm{p}\|$), velocity ($\|\delta\bm{v}\|$) and orientation ($\|\delta\bm{\theta}\|$) errors norms of the three controllers, as a function of time}
    \label{fig:all-errors}
\end{figure}

Fig.~\ref{fig:all-errors} compares the errors of all three controllers as a function of time during the real quadcopter's circular flight. In the following, we draw some conclusions about the performance of each controller during flight.

\textbf{Default QDrone Controller}:
The orientation error was noisy, averaging $10^\circ$ and peaking at $20^\circ$ around $25$~s. The velocity error remained a little above $0.05$~m/s, while the position error was about $0.2$~m, occasionally reaching $0.3$~m.


\textbf{Baseline SE$_2$(3) LQR Controller}:
Despite tuning limitations and the absence of a drag model, the LQR outperformed the default QDrone controller. When comparing Figs.~\ref{fig:xyz-lqr-exp} and~\ref{fig:xyz-quanser}, the LQR followed the desired trajectory more efficiently in the $X$ and $Y$ axes. 

Fig.~\ref{fig:all-errors} shows that the LQR achieved $\approx50\%$ lower position error and consistently maintained orientation error below $10\%$, with a less noisy response. In contrast, the Default QDrone Controller's orientation error exceeded $10^\circ$ for most of the time and was noisier. The velocity error was below $0.05$~m/s with the SE$_2$(3) LQR, slightly outperforming the Default architecture, which kept the error above this value.


\textbf{SE$_2$(3) MPC controller}: 
A comparison between Fig.~\ref{fig:xyz-lqr-exp} and Fig.~\ref{fig:xyz-mpc-exp} demonstrates that both MPC and LQR showed similar trajectory tracking in the $X$ and $Y$ axes, but MPC outperformed LQR  in $Z$-tracking, despite model mismatch and the absence of augmented state.

Overall, the MPC outperformed the Default QDrone controller and the LQR. As shown in Fig.~\ref{fig:all-errors}, the MPC's orientation error was approximately $5\times$ smaller than the QDrone's and $4\times$ smaller than the LQR's. The velocity error behaved similarly to that of the LQR, which outperformed Quanser's architecture. 
The position error with the MPC was $\approx4\times$ smaller than that of the Qdrone and almost $2\times$ smaller than that of the LQR.

The superior performance of MPC compared to LQR with the real quadcopter stems from its constrained optimization, which enabled smoother controller transitions at the beginning of the trajectory, allowing for better tuning of the gains. 

\section{Conclusion}
\label{sec:conclusion}
We implemented a control system architecture for real quadcopters modeled using the Lie group SE$_2$(3), and analyzed its performance under real-world conditions, including noise and unmodelled dynamics, compared to simulation. 

Additionally, we evaluated the architecture using two optimal control techniques, LQR and MPC, both in simulation and on a real-quadcopter. Results show that the SE$_2$(3) MPC outperforms the original SE$_2$(3) LQR and a commercial control system, achieving only $5\%$ tracking error compared to $20\%$ in the commercial solution.  

\balance

\bibliographystyle{IEEEtran}
\bibliography{References}

\end{document}